\documentclass[letterpaper, 10 pt, conference]{ieeeconf}  

\IEEEoverridecommandlockouts                              
\usepackage{graphics} 
\usepackage{epsfig} 
\usepackage{amsmath} 
\usepackage{amssymb}  

\usepackage{url}
\usepackage[ruled, vlined, linesnumbered]{algorithm2e}
\usepackage{verbatim}
\usepackage{soul, color}
\usepackage{lmodern}
\usepackage{fancyhdr}
\usepackage[utf8]{inputenc}
\usepackage{fourier}
\usepackage{array}
\usepackage{makecell}

\SetNlSty{large}{}{:}

\makeatletter

\newcommand{\Rom}[1]{\expandafter\@slowromancap\romannumeral #1@}
\makeatother

\title{\LARGE \bf
Learning Saliency Prediction From Sparse Fixation Pixel Map
}

\author{Shanghua Xiao
\\ Department of Computer Science, Sichuan University, Sichuan Province, China \\
{\tt 2015223040036@stu.scu.edu.cn}
}

\begin{document}

\maketitle
\thispagestyle{plain}
\pagestyle{plain}

\begin{abstract}
Ground truth for saliency prediction datasets consists of two types of map data: fixation pixel map which records the human eye movements on sample images, and fixation blob map generated by performing gaussian blurring on the corresponding fixation pixel map. Current saliency approaches perform prediction by directly pixel-wise regressing the input image into saliency map with fixation blob as ground truth, yet learning saliency from fixation pixel map is not explored. In this work, we propose a first-of-its-kind approach of learning saliency prediction from sparse fixation pixel map, and a novel loss function for training from such sparse fixation. We utilize clustering to extract sparse fixation pixel from the raw fixation pixel map, and add a max-pooling transformation on the output to avoid false penalty between sparse outputs and labels caused by nearby but non-overlapping saliency pixels when calculating loss. This approach provides a novel perspective for achieving saliency prediction. We evaluate our approach over multiple benchmark datasets, and achieve competitive performance in terms of multiple metrics comparing with state-of-the-art saliency methods.
\end{abstract}

\section{INTRODUCTION}

\par Saliency prediction aims to extract a mask map, termed "saliency map", that gives the probability of most attractive part from a given image. Such a saliency map can be utilized for multiple applications, such as image thumbnailing \cite{b1} and content-aware image resizing \cite{b2}. By reducing the scale of the field-to-be-perceived to specific salient region, it can also be used as preprocessing method to further speed-up other visual tasks, e.g. \cite{b3,b4}. The extensive usages leads to the popularity of saliency prediction, and recently many approaches are proposed towards this issue.

\par Essentially, the task is to map the implicated saliency information from the high-dimensional image data to low-dimensional saliency map. This mapping usually takes the form of extracting a gray scale map of salient area indicated by bright gaussian blob from original image. Traditional approaches hierarchically model the handcrafted features to extract saliency in unsupervised style \cite{b5}. More recently, deep neural networks (DNNs) based methods such as Deep Fixation \cite{b6} and SALICON \cite{b7} improved saliency prediction performance by a great margin, especially by transferring the pretrained model with rich semantic features learned from large scale image classification task to boost up the learning. Saliency prediction datasets usually have two types of labels: fixation pixel map which records the human eye movements in discrete individual pixels, and fixation blob map generated by performing gaussian blurring on the corresponding fixation pixel map. Current approaches tend to learn saliency prediction by regressing the input image to a gray scale map with fixation blob as label. As saliency prediction field continuously developing, many new approaches have been proposed to perform saliency prediction in such fixation blob regression style, yet learning saliency prediction from the raw format of fixation pixels is not been explored.


\par Here in this work, we propose a novel approach of learning saliency prediction from fixation pixel map instead of fixation blob map. We use clustering to construct a sparse fixation pixel maps label set from the raw fixation pixel maps.  When applying naive regression loss functions such as Kullback-Leibler Divergence and Mean Squared Error on such sparse output and label, nearby but not overlapping saliency pixels will cause false penalty and lead to undesirable results. Thus we propose a novel loss function with max-pooling transform on output to learn from such sparse fixation pixels.

\begin{figure}[!t]
\centering
    \includegraphics[width=3.3in]{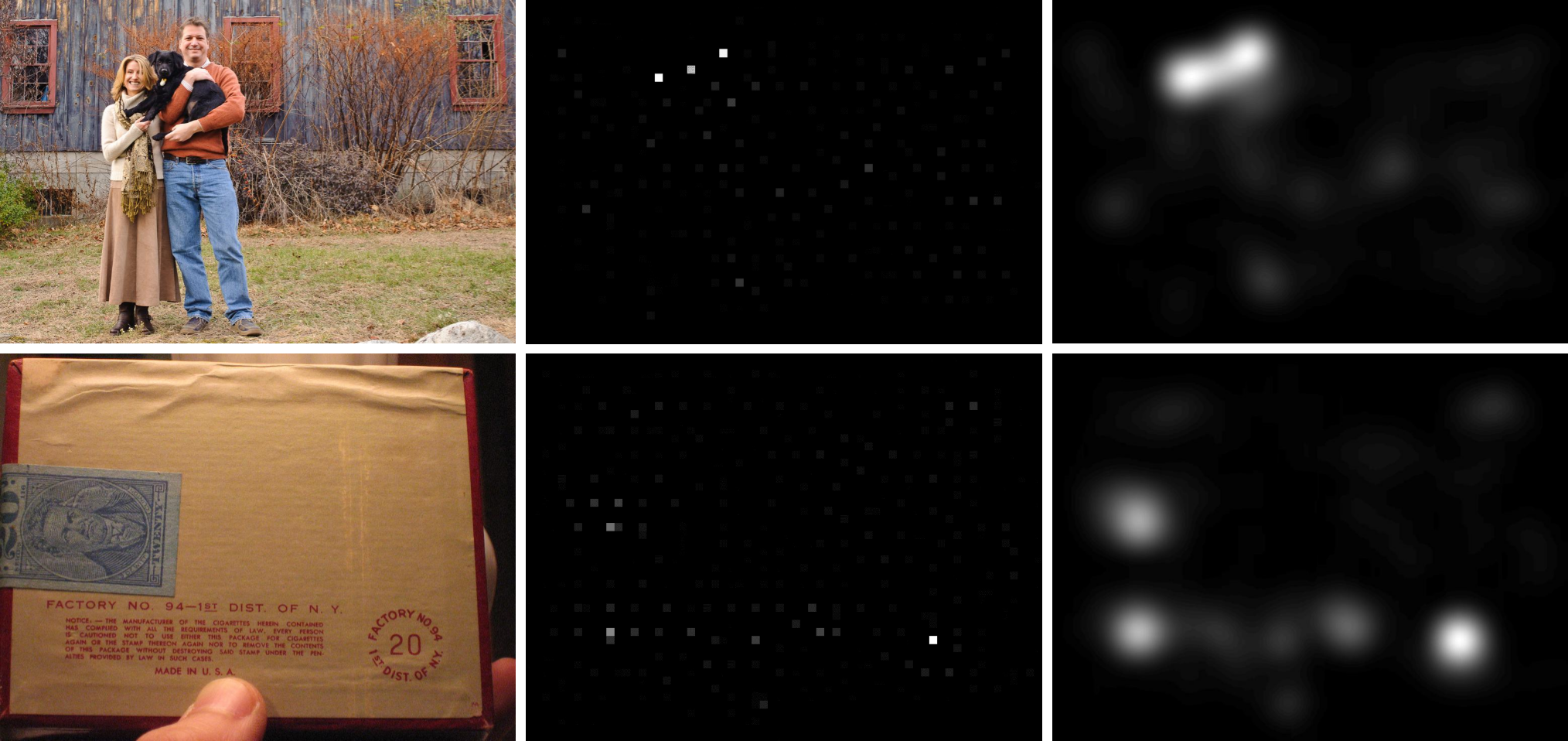}
    {Figure 1: Sample image with sparse fixation pixel output and gaussian blurred saliency map.\label{fig1}}
\end{figure}

\par We summarise the contribution of our work as follows:

\begin{itemize}
\item a first-of-its-kind approach of learning saliency prediction from sparse fixation pixels instead of fixation maps.

\item a novel loss function for training from such sparse label fixation map.
\end{itemize}

\section{Relative Works}

\par Current saliency prediction approaches which directly learn from fixation blob map can be organized into two categories: traditional approach which models handcrafted visual cues and DNNs based approach which models automatically learned visual features.

\begin{figure*}[!t]
\centering
    \includegraphics[width=7.0in]{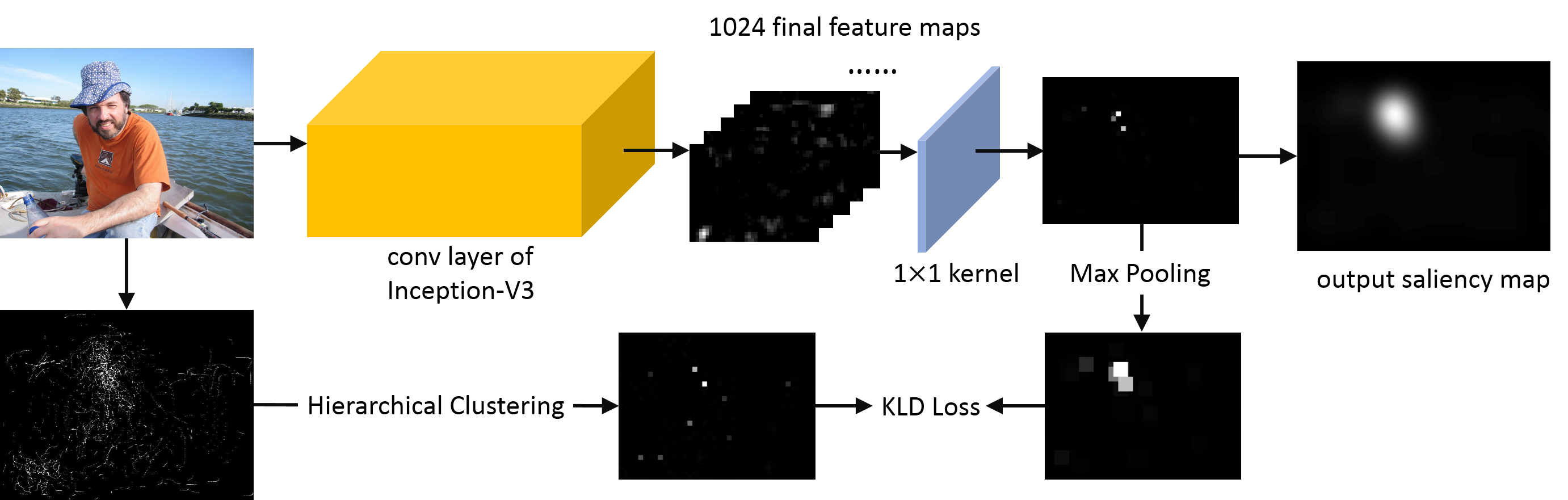}
    {Figure 2: Pipeline of training the proposed approach.\label{fig2}}
\end{figure*}

\par Traditional saliency prediction approaches are mainly driven by biological and psychological studies on human attention mechanism. Following the early study "Feature Integration Theory" \cite{b8} on human visual attention mechanism, Koch et al. propose a biological plausible visual attention model \cite{b9} which combines low level visual features such as color and contrast to produce a saliency indicating map. Later, Itti et al propose a saliency prediction approach based on the behaviour and neural architecture of primate visual system \cite{b10}. They extract low level visual cues into multiple conspicuity map, and use a dynamic neural network to integrate them into a single saliency map.

\par More recently, deep convolutional neural networks (CNNs) succeed in various computer vision tasks such as image classification \cite{b11,b12} and visual tracking \cite{b13,b14}, and became a common choice for modeling new saliency prediction approaches in data driven fashion. An early work of applying CNNs to saliency prediction is ensembles of Deep Networks (eDN) \cite{b15}, which models a 3 layers convolutional layers for feature extraction and a following SVM for salient classification. Later, Deep Gaze \cite{b16} adopt a deeper architecture with convolutional layers from AlexNet \cite{b17} to extract feature maps from different levels. The feature maps then integrated into saliency map by a learned linear model. As Deep Gaze introduces transfer learning into saliency prediction, later approaches extensively utilize trained models from image classification to boost up the saliency prediction performance. Kruthiventi et al. propose an fully convolutional neural network model named Deep Fix \cite{b6}, which utilize inception module and kernel with hole to extract multiple scale features, and applys VGG-16 \cite{b11} pretrained model to initialize its early feature extraion layers. In the work of SALICON \cite{b19}, Huang et al. explored multiple pretrained models from image classification for feature extraction in saliency prediction, including AlexNet, VGG-16 and GoogLeNet \cite{b20}.

\par Task-oriented loss functions are also explored to improve the saliency prediction performance. Normalized saliency map can be understood as a spatial probability distribution, and Saumya et al. proved that distributional perspective loss functions outperform standard regression loss funtions in their PDP model \cite{b21}. They explored multiple probabilistic distribution distance as loss function, namely $\chi^{2}$ Divergence, Total-variation Distance, Cosine Distance, Bhattacharyya Distance and Kullback-Leibler Divergence, and all achieved better performance than Euclidean and Huber distance. Evaluation metrics for saliency prediction are also explored as loss function in SALICON \cite{b7}. They utilize Normalized Scanpath Saliency (NSS), Similarity (Sim), Linear Correlation Coefficient (CC) and Kullback-Leibler Divergence (KLD) as loss function, and achieve best performance with KLD loss.

\par Different from previous works, we propose a saliency prediction approach which learn from sparse fixation pixel maps rather than fixation blob maps. We perform clustering on original fixation pixel map extract a sparse representation for each map, and fine tune a Inception-V3 model \cite{b22} to learn pixel level saliency information from the new label. Inspired by task-oriented losses, we apply KLD as loss function, and perform max-pooling on the output sparse saliency map to avoid false penalty on nearby but not overlapping saliency pixels between output and label maps.

\section{Proposed Approach}

\par In this section, we introduce our work of a novel approach for learning saliency prediction from sparse fixation pixel map. To learn from sparse fixation pixels, our work consists of two steps: constructing a new type of ground truth fixation map, and designing a DNN based saliency prediction model.

\par Human attention is mostly draw by certain objects, and saliency prediction also activates mostly on certain salient objects. CNNs use hierarchical integrating of visual features to percept different objects, which is eventually represented by strong activation on the center of the object region in the corresponding feature map. Thus besides regressing the probabilistic distribution map, saliency could also be learned from extracting salient level for certain objects in given image. In most saliency prediction datasets, the final ground truth fixation pixels is constructed by aggregating the fixation pixels from multiple observers. We explored raw fixation pixels as discrete distribution samples and find it roughly obey the mixed Gaussian distribution, thus it is reasonable to assume that salient object center is located in the center of corresponding fixation pixel cluster. Following this assumption, we can sparsely represent saliency map with activation center of all the salient objects in the stimuli, and construct such sparse activation ground truth by perform clustering on the raw fixation pixel map.

\par After constructing the sparse fixation label set, we use a deep convolutional neural network based model to predict sparse saliency pixels from a given image. Considering saliency prediction datasets are usually too small to train large scale DNNs from scratch, transfer learning is commonly applied to initialize network parameters from fine-trained classification models. We also build our model based on pretrained model, and the saliency prediction pipeline of our approach is shown in Figure 2.

\subsection{Sparse Fixation}

\par The fixation pixel ground truth from each datasets are usually distributed in a relatively uneven and clustered style, and each object is represented by multiple fixation pixels. The number of fixation pixels ranges from hundreds to thousands between different datasets, due to different extracting equipments and strategies. We assume pixels with gray scale greater than 250 being fixation points, then the average fixation pixel number is 66 for MIT1003\cite{b23}, 334 for CAT2000 \cite{b24}, and 4609 for SALICON \cite{b25}. For a salient object, the corresponding salient region is fill with random sampled fixation and non-fixation pixels in the fixation pixel map. When learning from the such fixation pixel ground truth, the non-fixation pixels in salient area will cause false penalty when calculating loss in training phase, thus making the model harder to train. To learn saliency prediction from fixation points, we need to sparsify fixation pixels to a level that roughly one fixation pixel represents one object, while maintaining the representative saliency information.

\par We perform clustering the raw fixation pixel map to extract sparse fixation pixels. We cluster the fixation points to certain amount of clusters, and use cluster center to represent each clusters. The cluster centers are calculated by the average pixel location in each cluster, and thus forming the sparse representation of fixation information.

\begin{figure}
\centering
    \includegraphics[width=3.3in]{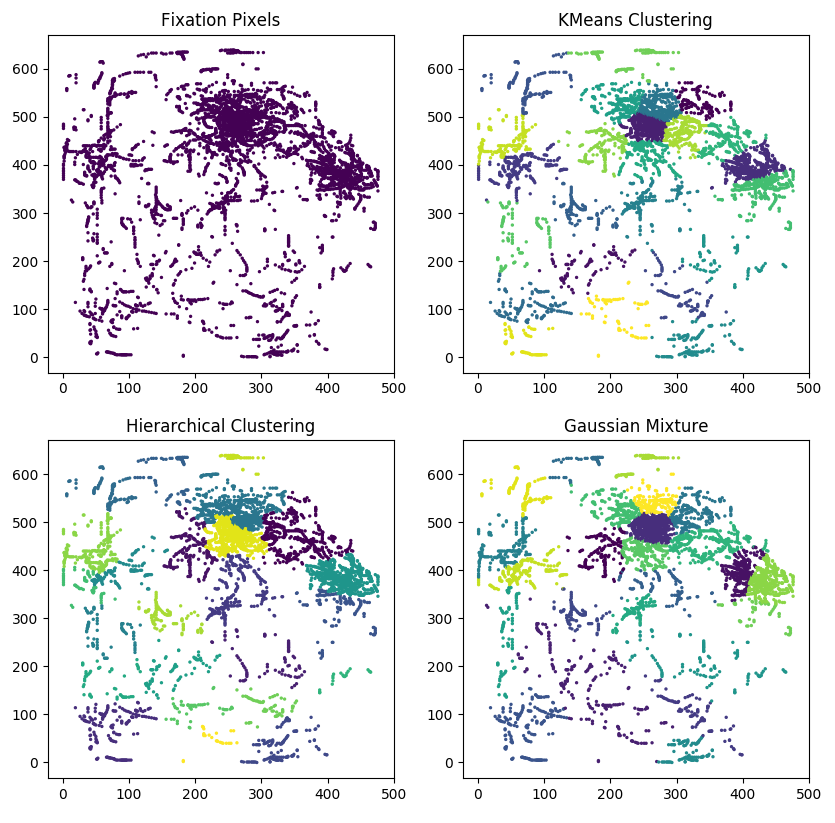}
    {Figure 3: Performance of different clustering methods on raw fixation pixel map from SALICON dataset\label{fig3}}
\end{figure}

\par To find an appropriate clustering method which maximize the fixation sparsity while preserving most salient information, we explored multiple clustering methods with various cluster number and parameter setups for fixation pixels sparsification. We visualize multiple results in Fig.3 to give intuitively comparison between multiple clustering method on raw fixation map. In Fig.3 a raw fixation map from SALICON dataset are clustered into 24 clusters by KMeans, Hierarchical Clustering \cite{b26} and Gaussian Mixture \cite{b27}. From the raw fixation map we can see there are two salient objects represented by two spots of fixation pixels, thus the ideal cluster would be two main clusters on each object and others and others with fewer fixation pixels. As shown, the fixation pixels in the two salient spots are clustered into 4 clusters by Hierarchical Clustering, while clustered into 8 clusters by KMeans and Gaussian Mixtures. Thus we choose Hierarchical Clustering to sparsify the fixation information.

\begin{figure*}[!t]
\centering
    \includegraphics[width=7.0in]{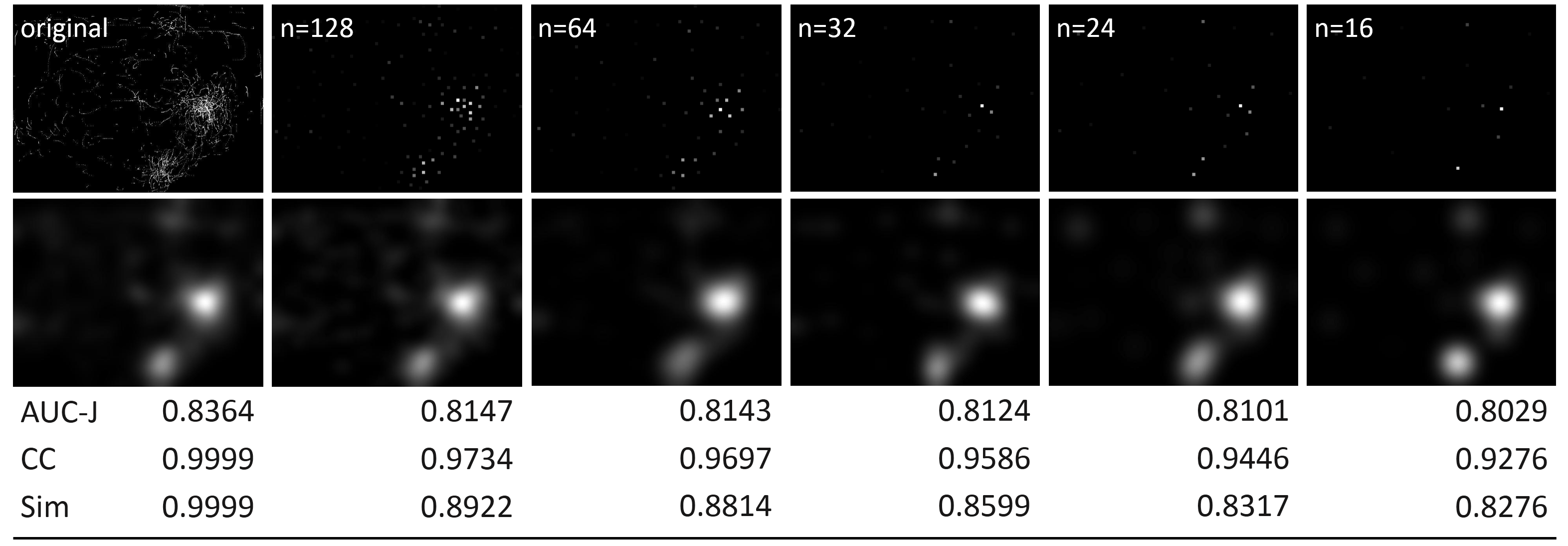}
    {Figure 4: Visualization of different cluster number\label{fig4}}
\end{figure*}

\par After select clustering for sparsification, we explored several cluster number to setup the clustering which maximize sparsity while appropriately preserve as much saliency information. The level of salient information preservation is evaluated by common metrics in saliency prediction such as AUD-Judd, NSS, Similarity and KL-Divergence in saliency prediction evaluation, and calculated between gaussian blurred raw fixation map and sparse fixation map. The result and visualization for different cluster number setup is shown in Fig.4. We select 24 to setup the cluster number for clustering on fixation pixel maps.

\par Finally, we construct a new ground truth dataset with 10000 sparse fixation map label from SALICON dataset. We use Hierarchical Clustering with cluster number set to 24, affinity set to Euclidean distance and linkage set to ward linkage.

\subsection{Network Architecture}

\par We model our approach using deep CNNs in fully convolutional fashion. We apply the convolutional layers of Inception-V3 \cite{b22} model to extract visual features from input images. The Inception-V3 model is trained on ImageNet \cite{b28} classification dataset with one million images, thus its kernels have a strong representative power. At the top of Inception-V3 convolutional layers, we add a simple 1$\times$1 kernel to reduce the final feature maps into a single output saliency map with detected sparse activated pixels.

\begin{figure}
\centering
    \includegraphics[width=3.3in]{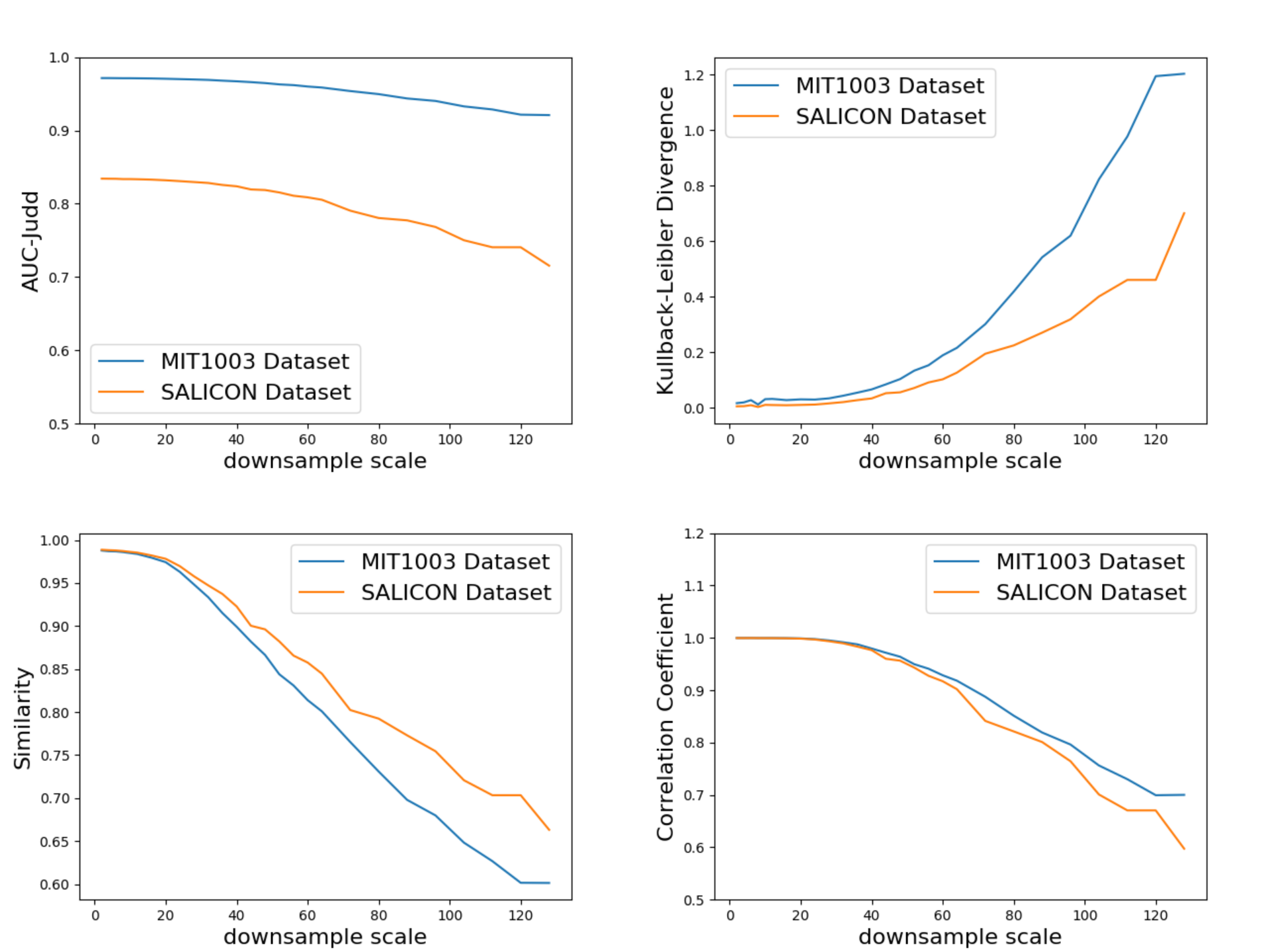}
    {Figure 5: Precision loss of different scale downsampled saliency maps on MIT1003 and SALICON dataset\label{fig5}}
\end{figure}

\par The convolutional part of Inception-V3 model consists of 23 convolution and pooling layers, in which 5 of the layers will cause 2-times downsampling, including 1 convolutional layer with stride of 2 and 4 pooling layers with stride of 2. In such structure the input image will be downsampled into the scale of 1/32 comparing with its original size, that is 20$\times$15 for image with size of 640$\times$480 from selected dataset SALICON. We evaluate the accuracy loss on different downsampled-upsampled fixation map with original fixation map to find an appropriate down sample scale. We evaluate the loss on SALICON and MIT1003 datasets and plot the loss in Fig.5. We can see that after 20-times downsample scale the accuracy loss drops fast. Since 32-times downsample by Inception-V3 model maybe too large for saliency prediction, we choose the downsample scale of 16 and modify the original network structure to achieve so. The higher layer of DNN has a strong dependency on previous layers, thus we modify the higher layer downsample scale to minimize the effect. We replace the stride of 2 in the last inception module by 1, and keep a downsample scale of 1/16.

\subsection{Pooling KLD Loss}

\par Since the saliency map is like discrete probabilistic distribution, we use Kullback-Leibler Divergence (KLD) as loss function. Let $label$ denotes the ground truth sparse fixation map, and $pred$ denotes the output sparse saliency map. The original KLD loss is in the form of

\begin{equation}KLD=\sum_{i=1}^{N}label_{i}*(log(label_{i})-log(pred_{i})).\label{eq}\end{equation}

where $label$ and $pred$ is a normalized N-dimensional distribution. The original form in KLD performs well on pixel-wised regression approach, but poorly on clustered center approach. KLD calculate the distance between $label$ and $pred$ by cumulating the pixel-wise difference on the two distribution matrix, thus two closely neighbored activated point from each matrix has no contribution to the closeness. This will cause false penalty between two spatially close but non-overlapping activated point from two matrix, and further results in wrong direction at gradient descending. To tackle this issue, we propose an alternative for original KLD loss, termed "pooling KLD".

\par Pooling KLD first extract a new output note by $pred_{p}$ by perform a padded max-pooling on original $pred$, then calculate the KLD on $label$ and $pred_{p}$. Note that we do not normalize the $pred_{p}$ again, since the $label$ in our case is sparse thus could perform as a gateway for cumulation, and the False Positive caused by max-pooling will be ignored if the corresponding gateway is not activated. The sparse label matrix and max pooling prediction matrix allow us to take neighboring information into account and learn the saliency points in more representative and robust fashion.

\par We feature two sparse fixation maps with nearby but non-overlapping activation in Fig.6, and calculate the original and pooling KLD between them. As shown, the KLD of original $label$ and $pred$ is 5.50577, while the KLD of $label$ and $pred^{p}$ is 0.74061. As pooling KLD can efficiently suppress the false penalty, we use pooling KLD for training saliency prediction from such sparse fixation maps.

\begin{figure}
\centering
    \includegraphics[width=3.3in]{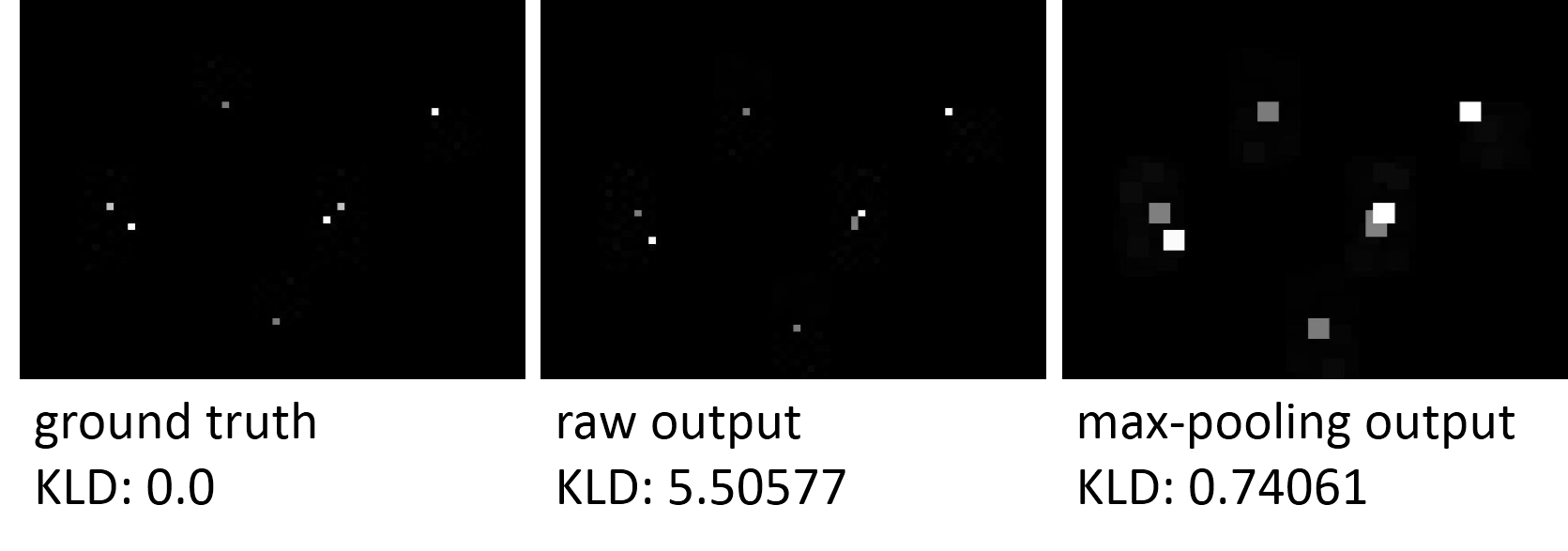}
    {Figure 6: Kullback-Leibler Divergence on raw fixation pixels and max-pooled fixation pixels with groud truth.\label{fig6}}
\end{figure}

\section{Experimental Result}

\subsection{Setups}

\par We construct the sparse fixation training set from the SALIency in CONtext (SALICON) dataset \cite{b19} by perform clustering with Scikit-Learn package \cite{b43}. SALICON is the largest open access dataset in the area of saliency prediction, which consist of 10000 training samples, 5000 validation samples and 5000 testing samples from MS COCO dataset \cite{b29}. The fixation information is gathered by an alternative eye tracking paradigm. Multiple observers use mouse to direct their fixation on image stimuli during 5 seconds free viewing and 2 seconds followed waiting interval, and the mouse trajectory is recorded and aggregated to indicate where people find most interesting in such stimuli. We perform clustering that we described previously to cluster fixation pixels of each sample into 24 clusters, and use the center of each clusters to represent the corresponding salient object.

\par At training phase, we use batch training to accelerate convergence and improve generalization capability, with batch size set to 16. We use Adam \cite{b30} optimizer to for fine-tuning the Inception-V3 model. Since saliency prediction and image classification can share the same low level features, we only fine-tune the final 6 inception blocks with higher level features by blocking the gradient at layer 6 from back propagating. The pretrained Inception-V3 model is downloaded from model zoo of MxNet framework. The learning rate for fine-tuning and the learning of final 1$\times$1 kernel is set to 0.00001. The entire training takes about 24 hours on a 12G ram NVIDIA Tesla K40m GPU with the MxNet deep learning framework \cite{b31} on Ubuntu 16.04 operation system.

\subsection{Evaluation Metrics}

\par At evaluation, multiple metrics are used, since previous study by Riche et al. \cite{b32} shows that no single metric has concrete guarantee of fair comparison. We briefly describe the used metrics for better understanding of the results. We denote $S$ for output saliency map, $G_{b}$ for ground truth fixation blob map and $G_{p}$ for ground truth fixation pixel map at following description.

\par \textbf{AUC}: Area Under ROC Curve (AUC) measures the area under the Receiver Operating Characteristic (ROC) curve, which consists of true and false positive rate under different binary classifier threshold between $S$ and $G_{p}$. Three different AUC implementations are mainly used in saliency prediction, namely AUC-Judd \cite{b32}, AUC-Borji \cite{b33} and shuffled-AUC \cite{b34}, and we mainly adopt AUC-Judd in the evaluation. They are differed in how the true and false positive rate are calculated. The higher the true positive rate and the lower the false positive rate are, the larger the AUC is, and thus the better performance we have.

\par \textbf{NSS}: Normalized Scanpath Saliency \cite{35} is the mean value at on the fixation pixels location in normalized $S$ with zero mean and unit standard deviation. Larger NSS score represents better performance.

\par \textbf{CC}: Correlation Coefficient (CC) measures the linear relationship between saliency matrix $S$ and $G_{b}$. CC score of 1 means $S$ and $G_{b}$ are identical, while 0 means $S$ and $G_{b}$ are uncorrelated. Thus larger CC score represents better performance.

\par \textbf{Sim}: Similarity (Sim) first normalizes $S$ and $G_{b}$ to $S^{N}$ and $G^{N}_{b}$, then calculate the sum of element-wised minimum between $S^{N}$ and $G^{N}_{b}$. Thus larger Similarity represent better performance.

\begin{table}
\caption{Evaluation on MIT300 Benchmark Dataset}
\label{table}
\begin{tabular}{|p{50pt}|p{30pt}|p{30pt}|p{20pt}|p{30pt}|p{30pt}|}
\hline
Models & AUC$\uparrow$ & SIM$\uparrow$ & CC$\uparrow$ & NSS$\uparrow$ & KL$\downarrow$ \\
\hline
our approach & 0.83 & 0.53 & 0.59 & 1.72 & 0.93 \\
DeepGaze*\cite{b18} & 0.88 & 0.46 & 0.52 & 1.29 & 0.96 \\
Deep Fix\cite{b6} & 0.87 & 0.67 & 0.78 & 2.26 & 0.63 \\
SALICON\cite{b7} & 0.87 & 0.60 & 0.74 & 2.12 & 0.54 \\
Deep Gaze\cite{b18}  & 0.84 & 0.43 & 0.45 & 1.16 & 1.04 \\
PDP\cite{b21} & 0.85 & 0.60 & 0.70 & 2.05 & 0.92 \\
BMS\cite{b37} & 0.83 & 0.51 & 0.55 & 1.41 & 0.81 \\
eDN\cite{b15} & 0.82 & 0.41 & 0.45 & 1.14 & 1.14 \\
Mr-CNN\cite{b38} & 0.79 & 0.48 & 0.48 & 1.37 & 1.08 \\
GBVS\cite{b39} & 0.81 & 0.48 & 0.48 & 1.24 & 0.87 \\
Judd\cite{b23} & 0.81 & 0.42 & 0.47 & 1.18 & 1.12 \\
LDS\cite{b40} & 0.81 & 0.52 & 0.52 & 1.36 & 1.05 \\
CAs\cite{b41} & 0.74 & 0.43 & 0.36 & 0.95 & 1.06 \\
\hline
\end{tabular}
\label{tab2}
\end{table}

\begin{table}\centering
\caption{Evaluation on CAT2000 Benchmark Dataset}
\label{table}
\begin{tabular}{|p{50pt}|p{30pt}|p{30pt}|p{20pt}|p{30pt}|p{30pt}|}
\hline
Models & AUC$\uparrow$ & SIM$\uparrow$ & CC$\uparrow$ & NSS$\uparrow$ & KL$\downarrow$ \\
\hline
our approach & 0.83 & 0.58 & 0.69 & 1.86 & 0.94 \\
SAM\cite{b42} & 0.88 & 0.77 & 0.90 & 2.40 & 0.69 \\
DeepFix\cite{b6} & 0.87 & 0.74 & 0.87 & 2.28 & 0.37 \\
eDN\cite{b15} & 0.85 & 0.52 & 0.54 & 1.30 & 0.97 \\
BMS\cite{b37} & 0.85 & 0.61 & 0.67 & 1.67 & 0.83 \\
Judd\cite{b23} & 0.84 & 0.46 & 0.54 & 1.30 & 0.94 \\
LDS\cite{b40} & 0.83 & 0.58 & 0.62 & 1.54 & 0.79 \\
GBVS\cite{b39} & 0.80 & 0.51 & 0.50 & 1.23 & 0.80 \\
CAs\cite{b41} & 0.77 & 0.50 & 0.42 & 1.07 & 1.04 \\
\hline
\end{tabular}
\label{tab3}
\end{table}

\par \textbf{KLD}: Kullback-Leibler Divergence (KLD) is a non-symmetric metric. It measures the information lost when using $S$ to encode $G_{b}$. Lesser KLD score represents better saliency prediction performance.

\subsection{Result}

\par We evaluate our approach on multiple benchmark datasets, and the results are as follows:


\par \textbf{MIT300} We mainly evaluate our model on the testing set of MIT300 \cite{b36} benchmark dataset. The MIT300 benchmark dataset is composed of 300 samples with various indoor and outdoor scenes and objects. The fixation information is extracted by directly recording the eye movements of 39 observers at 3 seconds free viewing at given sample. To avoid overfitting the dataset, the ground truth fixation maps are held out at the benchmark server for evaluation remotely, and the maximum submission is limited to 2 times per month. The sample sizes from MIT300 are ranged with x-axis from 679 to 1024 and y-axis from 457 to 1024, which are larger than from SALICON that we train our model on. Thus when evaluating on MIT300 dataset, we first resize the sample with short axis to 480 and long axis accordingly. The evaluation results are show in Table.1.

\par \textbf{CAT2000} We also evaluate our approach on CAT2000 \cite{b24} benchmark dataset. The CAT2000 dataset consists of one training set with accessable ground truth and one testing set with held out ground truth fixation maps. The training and testing set contains 20 different categories (100 images for each one) from \emph{Action} to \emph{Line Drawing}. The fixations are integrated from 5 seconds free viewing of 24 observers. Since the sample size of CAT2000 dataset is 1920$\times$1080, we resize the samples to 854$\times$480 for evaluation.

\par The evaluation in both datasets shows the practicability of learning saliency prediction from fixation pixels.

\begin{figure*}
\centering
    \includegraphics[width=7.2in]{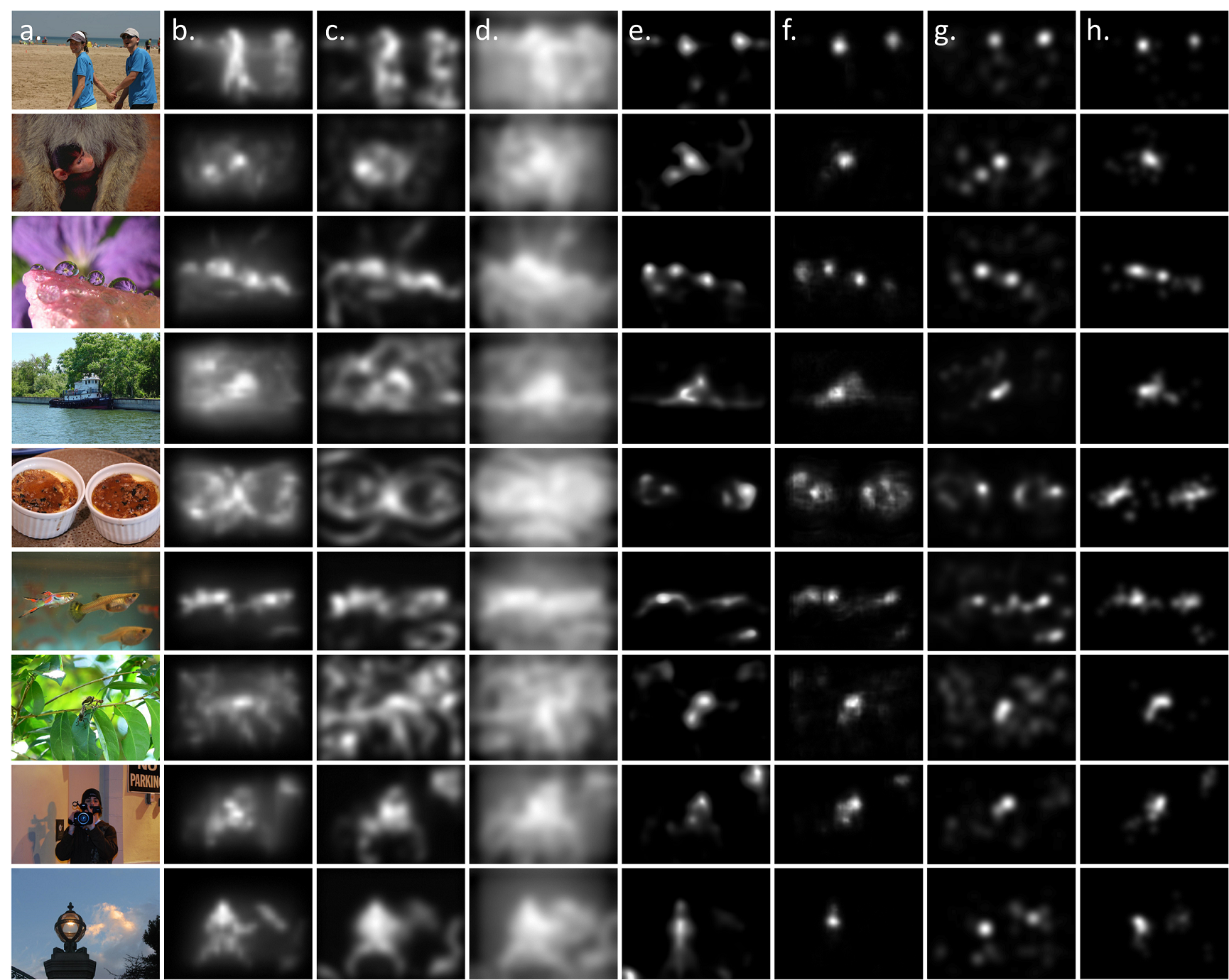}
    {Figure 7: Comparison on MIT1003 benchmark dataset. The columns from left to right are: a. input sample, b. GBVS \cite{b39}, c.Itti \cite{b44}, d. eDN \cite{b15}, e. DeepGaze II \cite{b18}, f. ML-NET\cite{b45}, g. our approach and h. ground truth.
    \label{fig7}}
\end{figure*}

\section{Conclusion}

\par In this work, we propose a first-of-its-kind method of learning saliency prediction from sparse fixation pixel map instead of gaussian blurred fixation map. A sparse fixation pixel map is extracted by hierarchical clustering the raw fixation ground truth and use the cluster center and sample number to represent the location and salient level of corresponding object. To tackle the problem of false penalty in sparse fixation regression, we propose a novel loss function with max pooling on the output. The proposed approach achieves state-of-the-art performance in multiple benchmark datasets, and provide a novel perspective on how saliency prediction can be learned.

\end{document}